\documentclass{article}
\usepackage{spconf,amsmath,graphicx}
\usepackage{subcaption}
\usepackage{multirow}
\usepackage{booktabs}
\usepackage{graphicx}
\usepackage{xcolor}
\usepackage{eqparbox}
\usepackage{arydshln}
\usepackage{url}

\usepackage[inline]{enumitem}

\newcommand{\comment}[1]{}

\title{An Analysis of Semantically-Aligned Speech-Text Embeddings}
\name{Muhammad Huzaifah, Ivan Kukanov}  
\address{
  Institute for Infocomm Research,
  Agency for Science, Technology and Research, Singapore
  }
\begin{document}
%

\maketitle

\begin{abstract}

Embeddings play an important role in end-to-end solutions for multi-modal language processing problems. Although there has been some effort to understand the properties of single-modality embedding spaces, particularly that of text, their cross-modal counterparts are less understood. In this work, we study some intrinsic properties of a joint speech-text embedding space, constructed by minimizing the distance between paired utterance and transcription inputs in a teacher-student model setup, that are informative for several prominent use cases. We found that incorporating automatic speech recognition through both pretraining and multitask scenarios aid semantic alignment significantly, resulting in more tightly coupled embeddings. To analyse cross-modal embeddings we utilise a quantitative retrieval accuracy metric for semantic alignment, zero-shot classification for generalisability, and probing of the encoders to observe the extent of knowledge transfer from one modality to another.

\end{abstract}

\section{Introduction}
\label{sec:intro}

Dense vector representations are ubiquitous as inputs into deep learning models. In natural language processing (NLP), there is a significant body of work pertaining to methods that map high-level discrete entities such as words to a low-dimensional continuous space. These learned embeddings often contain useful properties including semantic regularity, that is words that have similar meaning appear closer to each other in vector space \cite{mikolov2013distributed}. More recent deep-learning approaches such as ELMo \cite{peters1luke} and BERT \cite{devlin-etal-2019-bert} can further account for context, enabling for example, the disambiguation of homonyms appearing in different sentences. The masked language modeling objective popularised by BERT has since been extended to the audio domain to learn contextualized speech representations directly from audio \cite{baevski2020wav2vec, hsu2021hubert}. Utilising such embeddings have been shown to improve performance across a broad range of downstream tasks for both speech and text, including spoken language understanding (SLU), automatic speech recognition (ASR), and question-answering (Q-A).

Given that many modern NLP problems such as the above are multimodal in nature, there have been attempts at unifying the embedding spaces of speech and text. The use-cases of joint speech-text embeddings can be broadly classified into three non-mutually exclusive categories: 
\begin {enumerate*} [label=\itshape\alph*\upshape)]
\item when intending to leverage data from both modalities as input into a single model, such as for multimodal translation \cite{zheng2021fused} \item when learning a semantic alignment between text and speech \cite{duquenne2021multimodal,chung2018unsupervised} that can be useful for data mining or retrieval \item when transferring knowledge encoded in a pretrained model from one modality to another or incorporating knowledge from both modalities, often seen in end-to-end solutions for speech translation (ST) \cite{ye2021end,han2021learning,dong2021listen,Tang2021IST}, SLU \cite{Denisov_2020,chung2020splat} or Q-A \cite{chuang2019speechbert}), that combine semantics from text with acoustics from speech
\end {enumerate*}. 

Several studies have been carried out to better understand learned embeddings derived exclusively from either text or speech, via linear probing \cite{conneau-etal-2018-cram,yossisenemb,47786}, geometrical analysis of the representation space \cite{ethayarajh2019}, or other intrinsic measures \cite{Pasad2021LayerWiseAO, schnabel2015evaluation}. Cross-modal embeddings on the other hand have mostly been evaluated only in the context of a downstream task, hence are not well-understood. We attempt to close this gap by studying the extent to which directly minimising a distance-based measure between speech and text inputs at sentence-level can produce joint embeddings that are desirable for the above use-cases. We employ a dual encoder arranged in a teacher-student setup, where the speech embedding space gradually adapts to the text embedding space. Variations of this training objective are found in the literature, either as a pretraining step in isolation \cite{duquenne2021multimodal,Denisov_2020,chung2020splat}, or optimised in conjunction with the downstream task \cite{dong2021listen,Tang2021IST}.    

Ensuing from the above use-cases, we analyse the following characteristics of the joint embedding:
\setlist{nolistsep}
\begin{itemize}[noitemsep]
    \item For one model to leverage both data modalities, respective representations with the same underlying semantics should be close in embedding space. We measure this using retrieval accuracy, comparing different training scenarios (pretraining and multitask training) against a strong baseline.
    \item The resultant speech-text alignment should be robust to new inputs. We further experiment on the generalisability of the cross-modal alignment through zero-shot classification of several speech datasets. 
    \item Knowledge transfer from one pretrained model to the other should occur via teacher-student learning. The extent of embedded knowledge present in the model before and after training was ascertained through extending the linear probing technique to cover both text and speech. 
\end{itemize}

\section{Methods}

\subsection{Joint embedding model overview}
\label{je_model}
\begin{figure}[!htb]
  \centering
  \includegraphics[width=\linewidth]{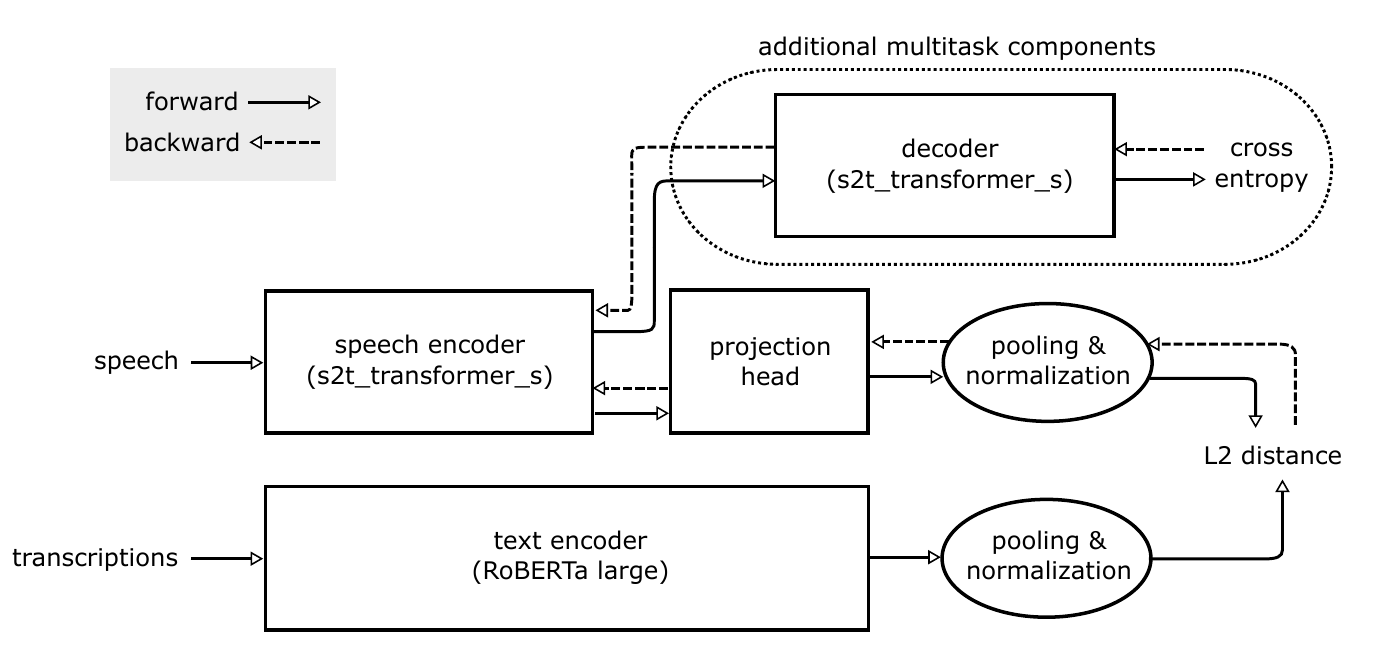}
  \caption{Architecture for joint speech-text embedding training. A decoder is added when training for the multitask objective.}
  \label{fig:model}
\end{figure}

To construct a joint speech-text embedding space, we train a model that learns to map a speech utterance and its corresponding transcription to the same point in embedding space, thereby coupling semantically-related objects from each domain. As illustrated in Fig.~\ref{fig:model}, the overall model contains two pipelines that respectively transform speech and text features derived from sentence-level utterance and transcription pairs into a hidden representation. The text pipeline consists of a pretrained language model acting as the teacher whereas a transformer-based speech encoder \cite{wang2020fairseq} acts as the student. A projection head block was utilised after the speech encoder to maintain consistent dimensions between the speech and text embeddings. The final output of each model was mean pooled over the length of the sentence, and the resultant fixed-length vector normalized to unit length. To bring the representations closer, we minimize the L2-distance between the outputs. Teacher-student learning occurs throughout the training by backpropagating the loss only through the speech pipeline while keeping the weights of the text model fixed. This progressively brings the speech embeddings closer to the space defined by the text encoder to eventually construct a joint representation space. 

To further enhance the representational strength of the speech embedding, we explored two techniques, namely pretraining and multitask training. For pretraining, the speech encoder was initialised with the encoder section of an encoder-decoder model trained for ASR. Meanwhile, in the multitask setup we train the speech encoder for ASR simultaneous to minimizing the distance between embeddings. Here, the overall loss is a combination of cross-entropy with label smoothing and the aforementioned L2-distance, weighted by $\gamma$ and $\beta$ respectively: \(L_{total} = \gamma L_{ce} + \beta L_{L2} \). The additional training schemes were found to be beneficial as preliminary experiments showed poor convergence for randomly-initialised models without extra supervision.


\subsection{Analysis methods}

\subsubsection{Retrieval accuracy}
The retrieval task relies on cross-modal alignment to fetch complimentary information and has found use in various settings including audio-text \cite{elizalde2019cross} and image-text \cite{pmlr-v139-jia21b}. We apply retrieval accuracy to assess the closeness of semantically-related speech and text embeddings in the joint space. Specifically, for a given speech input, we retrieve the text embedding with the highest cosine similarity (S$\rightarrow$T). If the retrieved text embedding belongs to its associated transcription then the prediction is counted. The converse is carried out to retrieve speech embeddings given a text input (T$\rightarrow$S).     

\subsubsection{Zero-shot utterance classification}
The joint embedding space was utilised for zero-shot classification as a way to study its potential for generalisation to unseen distributions. Here, the task is to correctly classify speech utterances according to a set of labels. Labels are treated as ``natural language'' rather than categorical abstractions and directly fed to the text encoder in the same fashion as the text transcriptions used in training. We then utilise the intrinsic alignment between the modalities to extract the closest label to a speech input, without having to further train a classifier over speech features. In practice, we find the speech-label embedding pair with the highest cosine similarity for each speech input. In this paper, we only explore datasets containing labels that are direct transcriptions of the speech input. Presumably, the information required to carry out tasks with a higher level of abstraction (such as emotion classification) would not have been effectively captured by the current setup and is left for future work.


\subsubsection{Probing for learned language features}
Probing was used to verify the kind of language features captured by the models and if any transfer learning from the text to the speech encoder had taken place during the joint embedding training. Given sentence embeddings produced by the frozen text encoder, we train several classifiers on different language tasks. If a classifier does well on a task, it implies that the embedding stores tangible information related to that task. To extend the technique to probe for transfer learning, we make the assumption that the embedding space of the speech encoder is already coupled to the text encoder through the earlier joint training and reuse the same classifiers to evaluate the speech embeddings. The same test sentences are transformed into speech using a text-to-speech (TTS) model prior to input into the speech encoder. Comparative analysis between the encoders before and after training is carried out.         


\section{Experimental Setup}

\begin{table*}[tb]
  \caption{Retrieval accuracy (\%) gauging how close speech-text embedding pairs are in the joint space. Models G and H refer to speech encoders in the randomly-initialised and pretrained settings prior to training, and serve as lowerbounds, whereas the cascade model I is an upperbound. Subsequent discussion will refer to the model variations by their designated letter.}
  \label{table:accuracy}
  \centering
  \begin{tabular}{@{} cr|cccccccc @{}}  
    \toprule
    \multirow{2}{*}{\textbf{Model}} & \multirow{2}{*}{\textbf{speech encoder \& training conditions}} & \multicolumn{2}{c}{\textbf{libri-clean}} & \multicolumn{2}{c}{\textbf{libri-other}} & \multicolumn{2}{c}{\textbf{MuST-C}} & \multicolumn{2}{c}{\textbf{CoVoST}} \\    
    
    & &\multicolumn{1}{c}{T$\rightarrow$S} & \multicolumn{1}{c}{S$\rightarrow$T} &     
     \multicolumn{1}{c}{T$\rightarrow$S} & \multicolumn{1}{c}{S$\rightarrow$T} & 
    \multicolumn{1}{c}{T$\rightarrow$S} & \multicolumn{1}{c}{S$\rightarrow$T} & 
    \multicolumn{1}{c}{T$\rightarrow$S} & \multicolumn{1}{c}{S$\rightarrow$T} \\
    \midrule
    A & random  & 30.23 & 12.02 & 15.89 & 5.78 & 1.97 & 0.90 & 0.01 & 0.01            \\
    B & pretrained & 98.82 & 83.42 & 94.25 & 67.88 & 44.33 & 15.55 & \textbf{2.15} & \textbf{0.35}                \\
    C & pretrained, only projection head updated & 98.32 & 63.89 & 91.15 & 47.64 & 37.90 & 10.60 & 1.62 & 0.15 \\
    D & random, multitask, $\gamma$=1, $\beta$=1 & 84.20 & 23.93 & 65.33 & 16.26 & 13.20 & 2.03 & 0.25 & 0.03    \\
    E & random, multitask, $\gamma$=1, $\beta$=10 & 88.82 & 29.01 & 72.07 & 20.01 & 17.43 & 3.10 & 0.49 & 0.02             \\
    F & pretrained, multitask, $\gamma$=1, $\beta$=100 & \textbf{99.12} & \textbf{91.91} & \textbf{95.75} & \textbf{79.58} & \textbf{50.71} & \textbf{21.04} & 1.55 & 0.25             \\
    \hdashline
    G & random, before JE training  & 0.04 & 0.00 & 0.00 & 0.07 & 0.06 & 0.06 & 0.01 & 0.01      \\
    H & pretrained, before JE training & 0.04 & 0.08 & 0.03 & 0.07 & 0.06 & 0.06 & 0.00 & 0.00   \\
    I & cascade & 99.47 & 99.50 & 97.03 & 97.62 & 62.40 & 65.01 & 3.26 & 4.75   \\
    \bottomrule
  \end{tabular}
\end{table*}

\subsection{Text and speech encoder architecture}
A pretrained RoBERTa large model \cite{liu2019roberta} from the Fairseq library \cite{ott2019fairseq} was adopted as the backbone of the text pipeline. RoBERTa improves on the masked language modeling objective of BERT \cite{devlin-etal-2019-bert} by tweaking several hyperparameters, removing the next-sentence prediction objective, and training on a much larger dataset, among other innovations. Before input, byte-level byte pair encoding (BBPE) was first applied to the text sequence. RoBERTa large contains 24 transformer layers and an output embedding size of 1024.

For the speech encoder, we used the \texttt{s2t\_transformer\_s} architecture from Fairseq, which consists of several convolutional neural network layers to downsample the input followed by 12 transformer layers, with an output embedding size of 256. The post-encoder projection head contains a Gaussian Error Linear Unit (GELU) activation between two linear layers, followed by dropout and layernorm. The speech pipeline for the multitask objective additionally includes a six layer transformer decoder. To ascertain the impact of pretraining, we compared randomly-initialised encoders against encoders pretrained for ASR in our experiments, for which the Fairseq checkpoint trained on LibriSpeech \cite{panayotov2015librispeech} was used. Raw audio was first converted into log-Mel filter bank spectrograms with 80 bins, frame length of 25ms, and shift of 10ms, then dynamically augmented following SpecAugment \cite{Park2019SpecAugmentAS} before input into the speech encoder.

\subsection{Joint speech-text embedding training}

The model was trained for the joint embedding objective on the LibriSpeech dataset using four NVIDIA V100 GPUs for 100 epochs with Adam optimizer, an inverse square root learning rate scheduler with a peak learning rate of $1\text{e-}3$, and 10000 warmup steps. For the multitask setup, early experiments showed that the contribution of the cross-entropy loss for ASR was about two orders magnitude larger than the L2 loss for embedding distance. The $\beta$ parameter was tweaked between 1-100 to increase the influence of the embedding distance. For all training variations, the checkpoint with the smallest validation loss was chosen for further analysis. In addition to the LibriSpeech test sets (libri-clean, libri-other), MuST-C (en-zh) \cite{di-gangi-etal-2019-must}, and CoVoST~2 (en) \cite{wang-etal-2020-covost} were also used for evaluation. For these, only the English speech segments and transcriptions were used.

\subsection{Baseline cascade model}
\label{baseline}
For a baseline, we directly matched text embeddings from RoBERTa against speech embeddings derived from a cascade model consisting of separate ASR and RoBERTa sections. The same LibriSpeech ASR checkpoint initialising the joint embedding model was used, first generating transcriptions from input speech that are then fed into RoBERTa to obtain the final embeddings. Unlike the end-to-end speech encoder section of the joint embedding model, it is not possible to jointly optimise the entire cascade model on further downstream tasks. Moreover, prosodic and acoustic information which may be useful for further tasks is lost in the ASR process. Nevertheless, this model provides a strong upperbound on how close speech and text embeddings can get since RoBERTa was utilised for both modalities. 

\subsection{Other analysis datasets}
We tested for zero-shot classification on three datasets, namely AudioMNIST \cite{becker2018interpreting}, Speech Commands \cite{warden2018speech}, and TIMIT \cite{garofolo1993timit}. AudioMNIST contains 30000 audio recordings of spoken digits 0-9 with 50 repetitions per digit per speaker. For labels, digits were spelled out instead of using the numerical form (i.e.\ ``one" rather than ``1"). A subset of Speech Commands was used comprising 10 auxiliary words: ``Bed", ``Bird", ``Cat", ``Dog", ``Happy", ``House", ``Marvin", ``Sheila", ``Tree", and ``Wow". Each word is spoken once by each speaker, making up 20408 recordings in total. The SX subset of TIMIT was used to explore joint embeddings with multi-word sentences acting as labels. SX sentences were designed to provide a good coverage of pairs of phones. We randomly selected 10 SX sentences among TIMIT's TEST set for labels, with seven pronunciations each for classification.

The text datasets provided by the SentEval toolkit \cite{conneau-etal-2018-cram} were used for probing. These span 10 semantic and syntactic tasks, from word-level objectives, such as word constituents, surface information like sentence length, to grammatical structures like past and present tense. We refer the reader to the original paper \cite{conneau-etal-2018-cram} for further details on each task. Each task is split into 100000 training, 10000 validation, and 10000 test examples. The classifier is a neural network comprising of two linear layers with dropout and a tanh activation. It was trained for 10 epochs and the checkpoint with the highest validation accuracy was chosen as the final model. The speech probing dataset was created by transforming the test sentences with an open-source TTS system by Silero \cite{Silero}, using the speaker model \texttt{lj\_16khz}.

\section{Results}

\subsection{How close are semantically-coupled speech and text embeddings in the joint space?}

\begin{figure}[htb]
  \centering
  \includegraphics[width=0.84\linewidth]{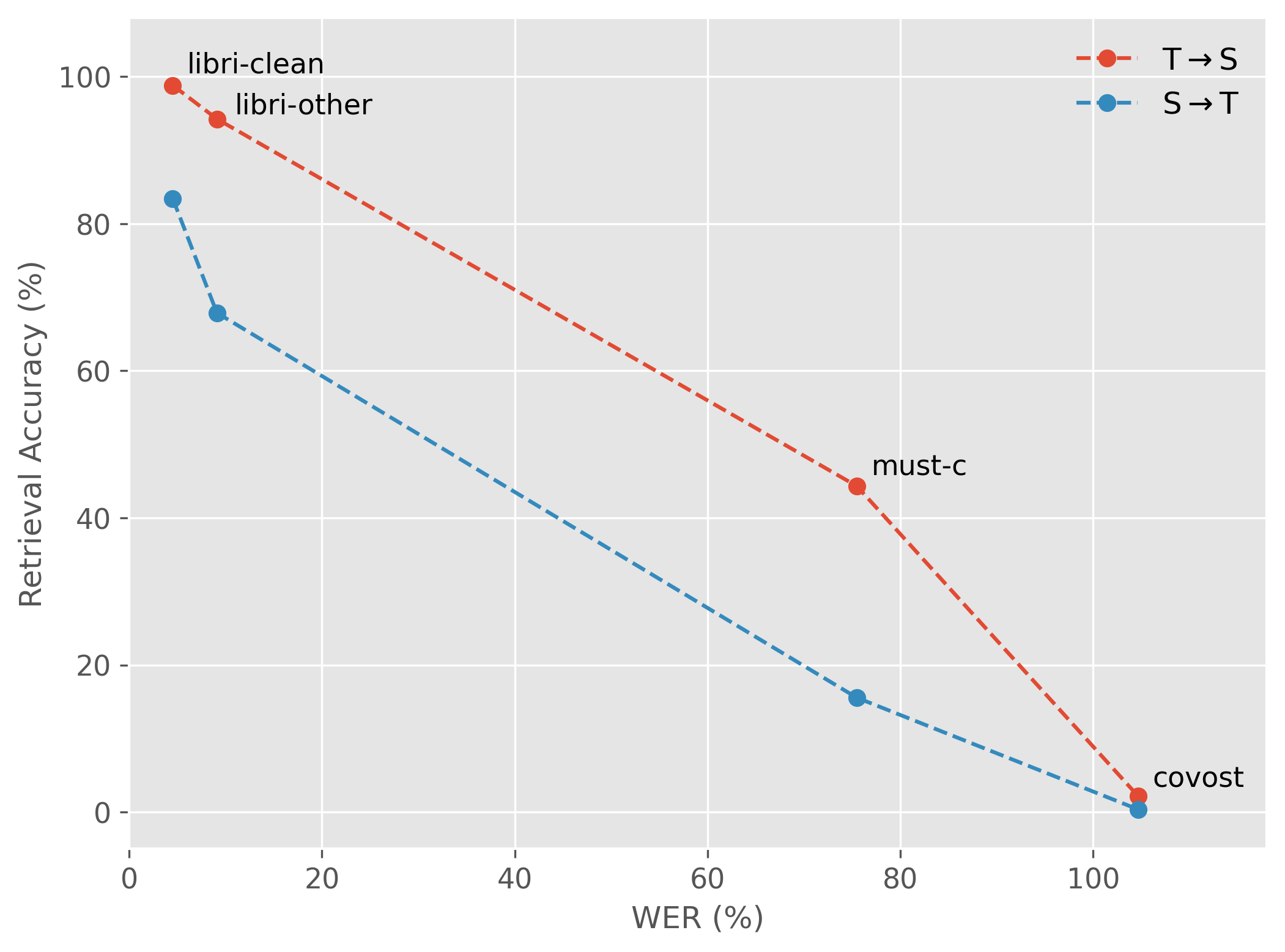}
  \caption{The effect of ASR pretraining on retrieval accuracy. Better ASR models (smaller WER) contributed to higher retrieval accuracy after the joint embedding training. WER was calculated on the original LibriSpeech pretrained checkpoint against the retrieval accuracy after training with this model, represented by Model B.}
  \label{fig:wer_acc}
\end{figure}

The retrieval performance across different initialisations of the speech encoder on various test sets is shown in Table \ref{table:accuracy}. The closeness of the resultant embeddings were highly affected by the training conditions. Prior to the joint training, speech and text embeddings were far apart irrespective of whether the speech encoder was pretrained for ASR (models G, H). ASR pretraining however significantly benefits the final closeness of the embeddings, comparing the randomly-initialised to the pretrained encoder after training (A, B). We believe given that the pretrained ASR encoder provides much better audio features from the outset, the training procedure better focuses on distribution shift rather than feature extraction. This is evidenced by the relatively slight drop in scores when freezing the speech encoder and only updating the projector head (C). Starting with a pretrained ASR model also provides better robustness on out-of-domain data, represented by MuST-C and CoVoST, although this is heavily influenced by the quality of ASR in relation to that dataset (Fig.~\ref{fig:wer_acc}). In general, better initial ASR performance, measured in terms of word error rate (WER), on a particular dataset led to closer cross-modal embeddings after training, as reflected in the retrieval accuracy. 

\comment{
 \begin{table}[htb]
   \caption{\todo{plotting figure} The effect of ASR model on retrieval accuracy S-T. The better the ASR model performance on evaluation dataset the better the semantic matching strength with text embeddings, the multitask\_asr model is from Table \ref{table:accuracy}, \emph{(pretrained, multitask, $\gamma$=1, $\beta$=100)}. Word error rate (WER,\%) / 100 - Acc.}
   \label{tab:wer_asr}
   \centering
   \begin{tabular}{@{} r|cccc @{}}
     \toprule
     \textbf{Models} & \textbf{libri-clean} & \textbf{libri-other} & \textbf{must-c} & \textbf{covost}\\
     \midrule
     multitask\_asr & 4.80 / 8.09 & 10.78 / 20.42 & 66.29 / 78.96 &	108.19 / 99.75\\
     \bottomrule
   \end{tabular}
 \end{table}
}

The multitask setups which train the speech encoder for ASR from scratch (D, E), did not perform as well as the models initialised with pretrained checkpoints (B, C, F), especially for S$\rightarrow$T retrieval accuracy. The learned speech features conferred by the ASR pretraining were found to be essential in the initial stages of the joint embedding training, leading to smoother optimization and closer final embeddings. The best overall model, model F, combines both pretraining and multitask, demonstrating the benefit of retaining ASR ability concurrent to minimizing the embedding distance. This end-to-end model came close to the performance of the baseline cascade model I, whose only source of error derives from imperfect ASR output, for in-domain T$\rightarrow$S retrieval but underperforms in the S$\rightarrow$T direction, especially on unseen data.

\begin{figure*}[tb]
     \centering
     \begin{subfigure}[b]{0.33\textwidth}
         \centering
         \includegraphics[width=\textwidth]{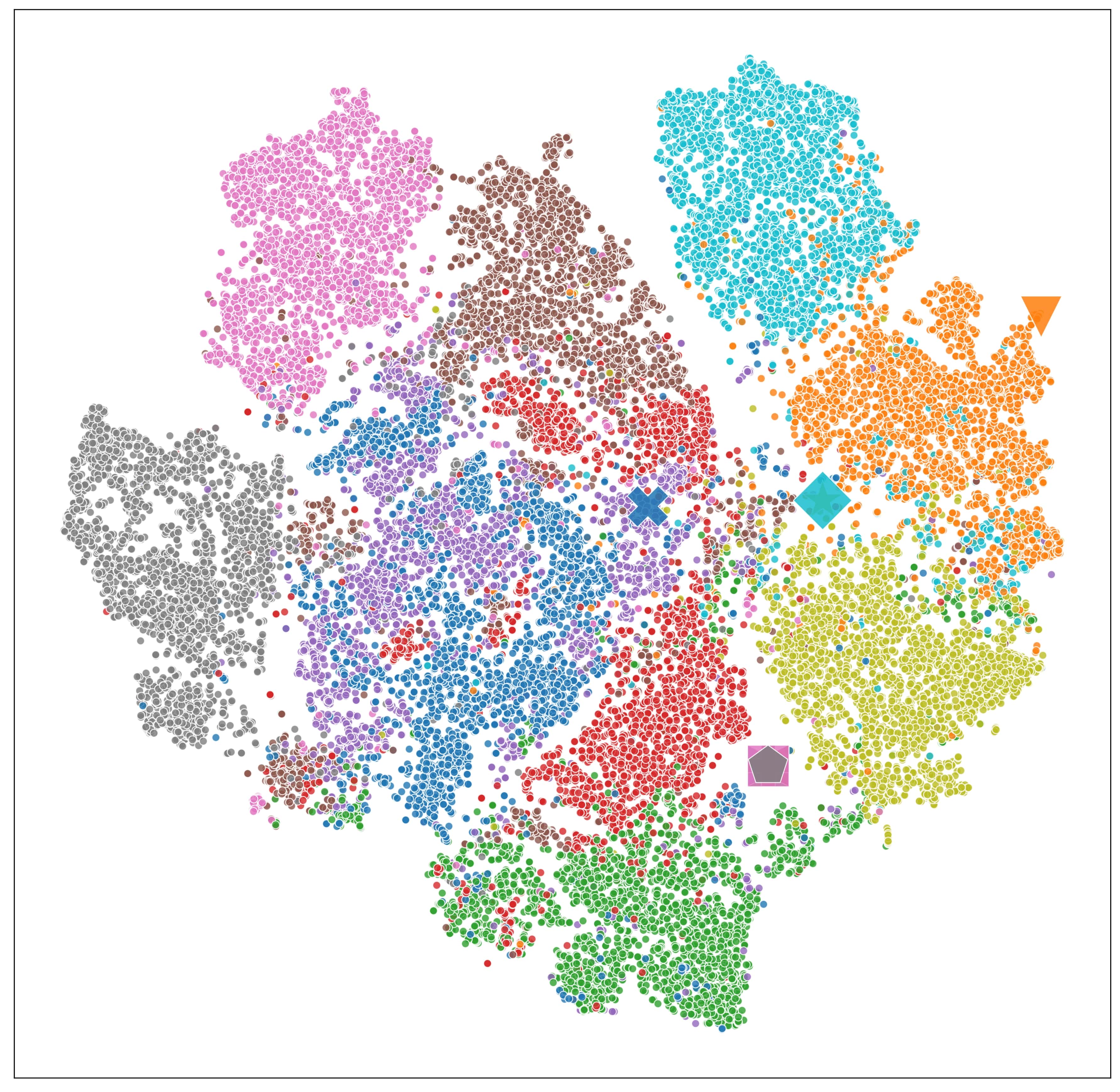}
         \label{fig:y equals x}
     \end{subfigure}
     \hfill
     \begin{subfigure}[b]{0.33\textwidth}
         \centering
         \includegraphics[width=\textwidth]{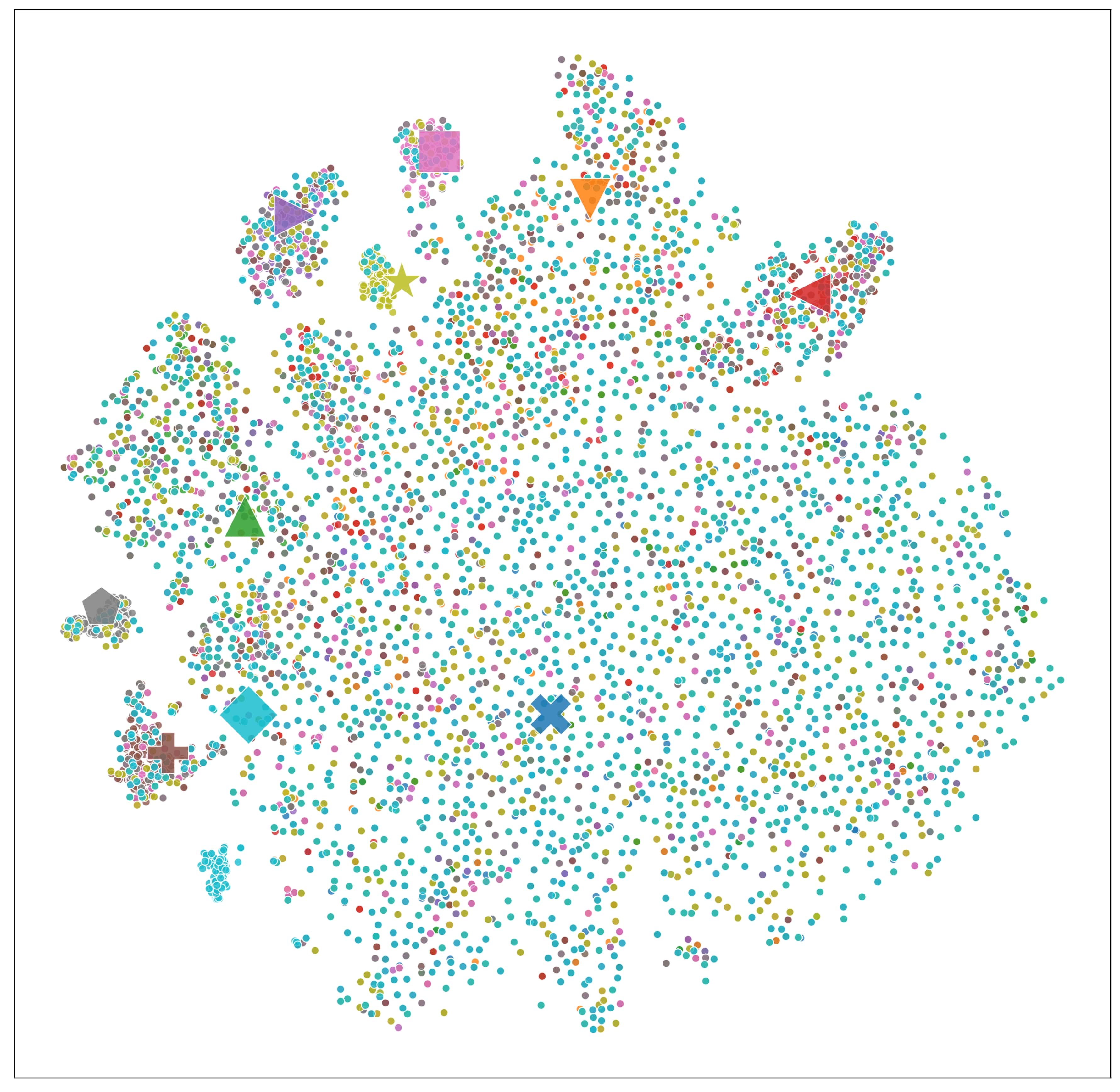}
         \label{fig:three sin x}
     \end{subfigure}
     \hfill
     \begin{subfigure}[b]{0.33\textwidth}
         \centering
         \includegraphics[width=\textwidth]{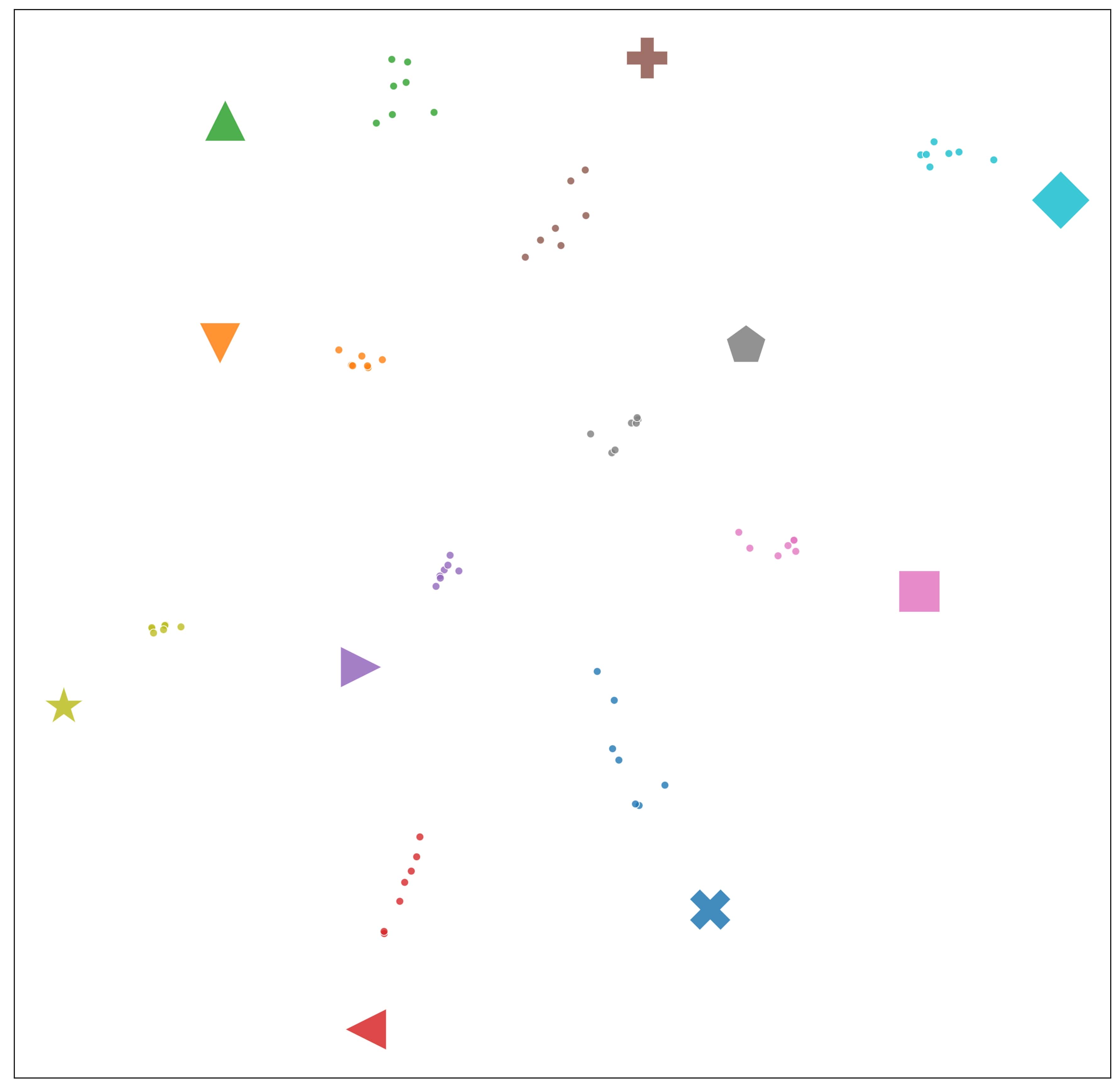}
         \label{fig:five over x}
     \end{subfigure}
        \caption{t-SNE plots of AudioMNIST (left), Speech Commands (centre), and TIMIT (right). Speech embeddings are marked by small circles while each label embedding is shown as a different, larger shape. Colour demarcates class for both modalities.}
        \label{fig:tsne}
\end{figure*}

It is noteworthy that the T$\rightarrow$S results were consistently higher than S$\rightarrow$T. We suspect this asymmetry stems from the teacher-student learning where the text embeddings remain static but the speech embeddings get pulled apart to fit the space. The final position of the speech embedding plane may leave a particular point closer to its text counterpart compared to other points on the speech plane and hence be retrieved correctly given a text input as in T$\rightarrow$S. Yet the same speech point may still be at a distance where it is closer to other unrelated points on the text embedding plane compared to its counterpart, making the corresponding S$\rightarrow$T retrieval wrong. This phenomenon is exacerbated in training setups with relatively poorer convergence, causing S$\rightarrow$T accuracy to deteriorate faster than T$\rightarrow$S in those cases. We note that such asymmetry was not reported in works without a teacher-student model such as ALIGN \cite{pmlr-v139-jia21b}, nor seen in the cascade model.

\subsection{Are the embeddings generalisable?}

While retrieval tests for a one-to-one match between a speech and text input, a more general task is to obtain a label for each input -- a many-to-one match. Table~\ref{tab:zeroshot} summarizes the zero-shot utterance classification results for the top performing models B and F on the previous retrieval metric, together with the untrained baseline model H. The accuracy, in particular on AudioMNIST and Speech Commands, were only slightly better than the baseline, which itself was close to random choice. However, scores were much higher on TIMIT. To provide more insight into the reasons behind this, speech and label embeddings were plotted using t-SNE in Fig.~\ref{fig:tsne}. 

\comment{
\begin{table}[htb]
  \caption{Classification accuracy (\%) of Models B, F, and H from Table \ref{table:accuracy} on AudioMNIST (AM), Speech Commands (SC) and TIMIT. Each dataset contained 10 classes.}
  \label{tab:zeroshot}
  \centering
  \begin{tabular}{@{} r|ccc @{}}
    \toprule
    \textbf{Dataset} & \multicolumn{1}{c}{\textbf{B}} & \multicolumn{1}{c}{\textbf{F}} & \multicolumn{1}{c}{\textbf{H}}\\
    \midrule
    AM & 12.59 & 15.88 &  10.28           \\
    SC & 14.47 & 14.27 & 9.31           \\
    TIMIT & 42.86 & 67.14 & 10.00              \\
    \bottomrule
  \end{tabular}
\end{table}
}

\begin{table}[htb]
  \caption{Classification accuracy (\%) of Models B, F, and H from Table \ref{table:accuracy} on AudioMNIST (AM), Speech Commands (SC) and TIMIT. Each dataset contained 10 classes.}
  \label{tab:zeroshot}
  \centering
  \begin{tabular}{@{} r|ccc @{}}
    \toprule
    \textbf{Model} & \multicolumn{1}{c}{\textbf{AM}} & \multicolumn{1}{c}{\textbf{SC}} & \multicolumn{1}{c}{\textbf{TIMIT}}\\
    \midrule
    B & 12.59 & 14.47 &  42.86           \\
    F & 15.88 & 14.27 & 67.14           \\
    H & 10.28 & 9.31 & 10.00              \\
    \bottomrule
  \end{tabular}
\end{table}

Distinct phenomena were observed for the three datasets. For AudioMNIST, speech embeddings showed clear separation among the different classes (Fig.~\ref{fig:tsne}, left). In contrast, several label embeddings were found to overlap in the t-SNE plot, suggesting that the RoBERTa text encoder was projecting them to a similar point in embedding space. This was a reasonable behaviour given that it was trained to produce contextualized word embeddings, and without any further context, individual digits may have been treated as semantically identical. Indeed, the more diverse set of words forming the Speech Commands labels were clearly separable (Fig.~\ref{fig:tsne}, centre). On this dataset however, the speech encoder did not produce embeddings aligned to the classes, outside of some minor clusters. This may be related to relatively poor multi-speaker ASR on short utterances, a setting far removed from the original ASR training, exacerbated by each speaker only providing a single utterance for each label. Since both encoders were trained on sentences rather than individual words, we expected them to do better on TIMIT, where each ``label'' was a full sentence (Fig.~\ref{fig:tsne}, right). Both clustering of the speech input and alignment between the two modalities were observed, corroborating the better accuracy scores. It is evident that a good alignment model must go hand in hand with a strong representation model for better generalisability in downstream speech-text tasks. 

\subsection{Is knowledge transfer taking place?}

To observe the extent of semantic knowledge transfer, we probed the RoBERTa-based text encoder, together with the speech encoders before (baseline H) and after joint embedding training (B and F), on 10 language tasks, shown in Table \ref{tab:prob}. The text encoder accuracy (RBT) provides an upperbound to the knowledge embedded in the teacher model. Notably, the text encoder itself did not perform well on several tasks, in particular ones that required word-level knowledge such as word content (WC) and semantic odd man out (SOMO). We attribute this to the mean pooling carried out post encoder, which may have resulted in more ambiguous word representations. Coordination inversion (CoordInv), where the order of coordinate clauses were inverted half the time, may also have been impacted by mean pooling the original embeddings.    

\comment{
\begin{table}[htb]
  \caption{Classification accuracy (\%) on 10 probing tasks for Models B, F, and H. RBT refers to the teacher text encoder model.
}
  \label{tab:prob}
  \centering
  \begin{tabular}{@{} r|c|cccc @{}}
    \toprule
    \textbf{Task} & \textbf{Classes} & \multicolumn{1}{c}{\textbf{RBT}} & \multicolumn{1}{c}{\textbf{B}} & \multicolumn{1}{c}{\textbf{F}} & \multicolumn{1}{c}{\textbf{H}}\\
    \midrule
    SentLen & 6 & 36.65 & 38.19 & 37.82 & 16.19             \\
    WC & 1000 & 0.10 & 0.10 & 0.00  & 0.10                \\
    TreeDepth & 7 & 20.48 & 19.60 & 19.95  & 18.47 \\
    TopConst & 20 & 29.25 & 22.68 & 21.57  & 5.14\\
    BShift & 2 & 74.82 & 54.19 & 54.42  & 48.83\\
    Tense & 2 & 81.89 & 75.41 & 77.59 & 50.01 \\
    SubjNum & 2& 78.58& 73.86& 74.95& 49.96\\
    ObjNum & 2& 76.26& 74.42& 75.66& 50.17\\
    SOMO & 2& 60.16& 49.78& 50.56& 49.66\\
    CoordInv & 2& 57.92& 50.45& 51.06& 51.19\\
    \bottomrule
  \end{tabular}
\end{table}
}

\begin{table}[htb]
  \caption{Classification accuracy (\%) on 10 probing tasks, with the respective number of classes inside each bracket. Change in accuracy after training is additionally provided for models B and F. RBT refers to the teacher text encoder. }
  \label{tab:prob}
  \centering
  \resizebox{0.995\linewidth}{!}{
  \begin{tabular}{@{} r|c|ccc @{}}
    \toprule
    \textbf{Task} & \multicolumn{1}{c}{\textbf{RBT}} & \multicolumn{1}{c}{\textbf{B}} & \multicolumn{1}{c}{\textbf{F}} & \multicolumn{1}{c}{\textbf{H}}\\
    \midrule
    SentLen (6) & 36.65 & 38.19 (+22.0) & 37.82 (+21.6) & 16.19             \\
    WC (1000) &  0.10 & 0.10 (+0.1) & 0.00 (-0.1) & 0.10                \\
    TreeDepth (7) & 20.48 & 19.60 (+1.1)& 19.95 (+1.5) & 18.47 \\
    TopConst (20) &  29.25 & 22.68 (+17.54)& 21.57 (+16.4) & 5.14\\
    BShift (2) &  74.82 & 54.19 (+5.4) & 54.42 (+5.6) & 48.83\\
    Tense (2)& 81.89 & 75.41 (+25.4) & 77.59 (+27.6) & 50.01 \\
    SubjNum (2)& 78.58& 73.86 (+23.9)& 74.95 (+25.0)& 49.96\\
    ObjNum (2)& 76.26& 74.42 (+24.3)& 75.66 (+25.5)& 50.17\\
    SOMO (2)& 60.16& 49.78 (+0.1) & 50.56 (+0.9)& 49.66\\
    CoordInv (2)&  57.92& 50.45 (-0.7) & 51.06 (-0.1) & 51.19\\
    \bottomrule
  \end{tabular}}
\end{table}

Overall, we observed evidence of knowledge transfer for some language properties but not all of them. This is characterised by the student speech encoders getting much closer to the teacher model's score after training compared to before, evident in tasks such as sentence length (SentLen), top constituents (TopConst) which tests for sentence structure, past and present tense (Tense), and plurality of the subject (SubjNum) or object (ObjNum) in a sentence. Some tasks that did not carry over may have been impacted by the continuous nature of audio in contrast to discrete text. For example, bigram shift (BShift), which inverts two adjacent words at random, may be much harder to detect directly from speech compared to text. Additionally, we acknowledge that some degree of error stems from the imperfect TTS conversion to create the speech probing dataset. For reference, the transformed SentLen test set had a WER of 26.78 with model F.  


\comment{

\subsection{Analysis in relation to ASR}
\label{sec:asr}

\note{probably remove this section}

We noticed since RoBERTa text model trained on the same domain as text corpus it is evaluated on (e.g., MuST-C), it benefits the ASR model trained in multitask manner with joint embeddings and outperform regular ASR model (transformer\_s), see Tab.~\ref{tab:wer_asr}. 

\begin{table}[htb]
  \caption{Comparing LibriSpeech ASR checkpoint \cite{panayotov2015librispeech} against multitask-trained ASR with joint embeddings from Tab.~\ref{table:accuracy} \emph{(pretrained, multitask, $\gamma$=1, $\beta$=100)}. Word error rate (WER,\%).}
  \label{tab:wer_asr}
  \centering
  \begin{tabular}{@{} r|cccc @{}}
    \toprule
    \textbf{Models} & \textbf{libri-clean} & \textbf{libri-other} & \textbf{must-c} & \textbf{covost}\\
    \midrule
    transformer\_s & 4.52 & 9.10 & 75.47 & 104.66  \\
    multitask\_asr & 4.80 &	10.78 &	66.29 &	108.19 \\
    \bottomrule
  \end{tabular}
\end{table}

}

\section{Conclusions}
We studied some intrinsic properties of a joint speech-text embedding space constructed through sentence-level semantic alignment between utterance-transcription pairs. The model was trained by minimizing the distance between the embedding outputs of a teacher RoBERTa text encoder and a student speech transformer encoder. We found that ASR pretraining of the speech encoder was essential for closer semantic alignment between the modalities, quantified in terms of retrieval accuracy. The performance could be further improved by combining it with a multitask objective incorporating both the joint embedding training and further ASR fine-tuning. The intrinsic alignment between the modalities may be potentially leveraged for more general tasks, such as utterance classification, even in the zero-shot setting. However, it was evident that each encoder had to capture the latent properties of the data to a suitable degree for this method to be viable, and that a good alignment model without a good representation model fails to generalise well. Linear probing of both encoders showed that not all linguistic properties are learned to the same degree by the student model during transfer learning, and that some operations such as pooling could potentially dilute specific language features such as word-level information. However, more experiments have to be carried out to understand the precise reasons behind why certain language features are transferred better than others and the potential impact on downstream tasks. In future work, we plan to also investigate embeddings and training schemes that would enable semantic coupling between less constrained speech-text pairs compared to utterances and transcriptions, that may be of interest to a wider range of downstream tasks.

\clearpage
\bibliographystyle{IEEEbib}
\bibliography{mybib}

\end{document}